\documentclass[conference]{IEEEtran}
\IEEEoverridecommandlockouts

\usepackage{cite}
\usepackage{amsmath,amssymb,amsfonts}
\usepackage{algorithmic}
\usepackage{graphicx}
\usepackage{textcomp}
\usepackage{xcolor}
\usepackage{booktabs}
\usepackage{tcolorbox}
\usepackage{float}
\usepackage{multirow}
\usepackage{pifont}
\usepackage{tabularx}
\usepackage{url}
\usepackage{hyperref}

\usepackage{lineno}

\def\BibTeX{{\rm B\kern-.05em{\sc i\kern-.025em b}\kern-.08em
    T\kern-.1667em\lower.7ex\hbox{E}\kern-.125emX}}

\begin{document}

\title{Making the Most of Limited Data: Score-Aware Training for Text-to-Music Generation}

\author{
\IEEEauthorblockN{Yun-Chen Cheng, Tzu-Hung Huang, Chih-Pin Tan}
\IEEEauthorblockA{
\textit{Graduate Institute of Communication Engineering, National Taiwan University, Taipei, Taiwan}\\
\{hhhhaura, zionhuang1107, tanchihpin0517\}@gmail.com}
}
\maketitle

\begin{abstract}
State-of-the-art text-to-music generation systems rely on massive proprietary 
datasets and industrial-scale compute, making it impossible to disentangle 
architectural contributions from resource advantages. We propose 
\textit{score-aware training}, which treats audio-caption alignment score as 
a direct supervision signal throughout the pipeline. Rather than discarding 
low-scoring segments, we repurpose them via a CLAP-conditioned Beta noise 
timestep schedule that routes them to high-noise training regimes, acting as 
an effective implicit regularizer. Complementarily, segment-level filtering 
removes the most misaligned examples, and a two-stage caption procedure 
bridges the distribution gap between verbose training captions and concise 
inference prompts. A REPA auxiliary loss further transfers structured semantic 
knowledge from pretrained CLAP and MuQ encoders without additional data. 
Our 450M-parameter FluxAudio-based system, submitted to the ICME 2026 
ATTM Grand Challenge Efficiency Track, ranked 2nd across both tracks in 
the objective evaluation and 3rd in the Efficiency Track in the final 
MOS evaluation.
\end{abstract}

\section{Introduction}
\label{sec:intro}

Text-to-music generation (TTM) has become a cornerstone of modern AI-driven music creation, powering commercial applications
that allow creators to produce music through natural language prompts alone. This shift toward text-driven interfaces reflects a broader democratization of music production, lowering the barrier for non-musicians to express musical ideas and enabling new forms of human-AI creative collaboration. Driven by advances in latent diffusion models, flow matching, and large-scale Transformer architectures, TTM systems such as MusicGen~\cite{copet2023musicgen} and Stable Audio Open~\cite{evans2025stableaudio} have achieved remarkable musical quality and semantic controllability. Yet state-of-the-art models are predominantly trained on massive proprietary 
datasets and industrial-scale computational infrastructure, creating a 
significant barrier for academic researchers who wish to study or reproduce 
these systems, let alone pursue algorithmic breakthroughs. Without 
controlled access to comparable data and compute, it becomes impossible to 
disentangle whether performance gaps stem from architectural choices or 
simply from differences in training resources~\cite{hsieh2026attm}.

The ICME 2026 Academic Text-to-Music (ATTM) Grand Challenge~\cite{hsieh2026attm} directly confronts this barrier by establishing a fair-play benchmark in which all participants must train generative models strictly from scratch on a standardized, CC-licensed subset of the MTG-Jamendo dataset~\cite{bogdanov2019mtg}. This controlled setting removes data scale and proprietary resources as confounding factors, enabling direct and reproducible comparison of algorithmic design choices. We participate in the \textbf{Efficiency Track}, which further imposes a strict upper bound of 500M parameters on the core generative model.

This setup raises a fundamental question: \textit{when data volume, data source, and model capacity are all fixed, what determines the ceiling of a model's performance?} We argue that the answer lies in \textit{how effectively} that data is used during training. In practice, large-scale music datasets such as MTG-Jamendo exhibit substantial heterogeneity in audio-caption alignment score: even within a single track, different segments vary widely in how faithfully they correspond to the associated caption, motivating fine-grained, segment-level score management throughout the training pipeline. Furthermore, this heterogeneity creates a fundamental trade-off between 
training data volume and alignment score: treating all segments uniformly risks 
polluting the training signal with misaligned examples, while aggressively 
filtering for only the highest-scoring segments sacrifices substantial 
amounts of potentially useful musical content.

These observations motivate a data-centric approach organized around the 
principle of \textbf{score-aware training}. We propose four complementary 
components: (i) \textbf{segment-level CLAP-guided filtering} to remove the 
most misaligned audio-caption pairs at sub-track granularity; (ii) a 
\textbf{CLAP-conditioned Beta noise timestep schedule} that repurposes 
lower-scoring segments by routing them to the high-noise training regime, 
where coarse content is useful and misalignment is less harmful~\cite{li2025baddata}; 
(iii) a \textbf{two-stage caption procedure} that fine-tunes on LLM-rewritten 
captions to bridge the verbose-training vs.\ concise-inference distribution gap; 
and (iv) a \textbf{REPA auxiliary loss}~\cite{yu2025repa} that transfers 
structured semantic knowledge from pretrained CLAP~\cite{elizalde2023clap} 
and MuQ~\cite{zhu2025muq} encoders without additional data.

Beyond data utilization, training a generative model from scratch on limited data poses a representation-learning challenge: the model must simultaneously discover the acoustic structure of music and align it with complex textual concepts. Yet structured semantic spaces capturing precisely these relationships have already been established by large-scale discriminative models. Rather than forcing our resource-constrained model to learn these abstractions in isolation, we incorporate a \textbf{representation alignment (REPA)}~\cite{yu2025repa} auxiliary loss that aligns the model's internal representations with pretrained embeddings from CLAP~\cite{elizalde2023clap} and MuQ~\cite{zhu2025muq}, transferring structured audio-semantic knowledge without additional data. In ablation, this yields a $+0.018$ CLAP score improvement and reduces FAD from 0.2856 to 0.2767.

Together, these four components demonstrate that careful handling of data quality and training dynamics can substantially advance text-to-music generation in the absence of industrial-scale resources.

\section{Methodology}
Our approach is organized around a unifying principle of \textit{score-aware training}: rather than treating all training data and all training steps equally, we systematically adapt each component of the pipeline---data selection, noise scheduling, text conditioning, and representation learning---based on an estimate of sample quality. We describe each component in turn below.

\subsection{Segment Filtering Pipeline}
\begin{figure}[htbp]
    \centerline{\includegraphics[width=\linewidth]{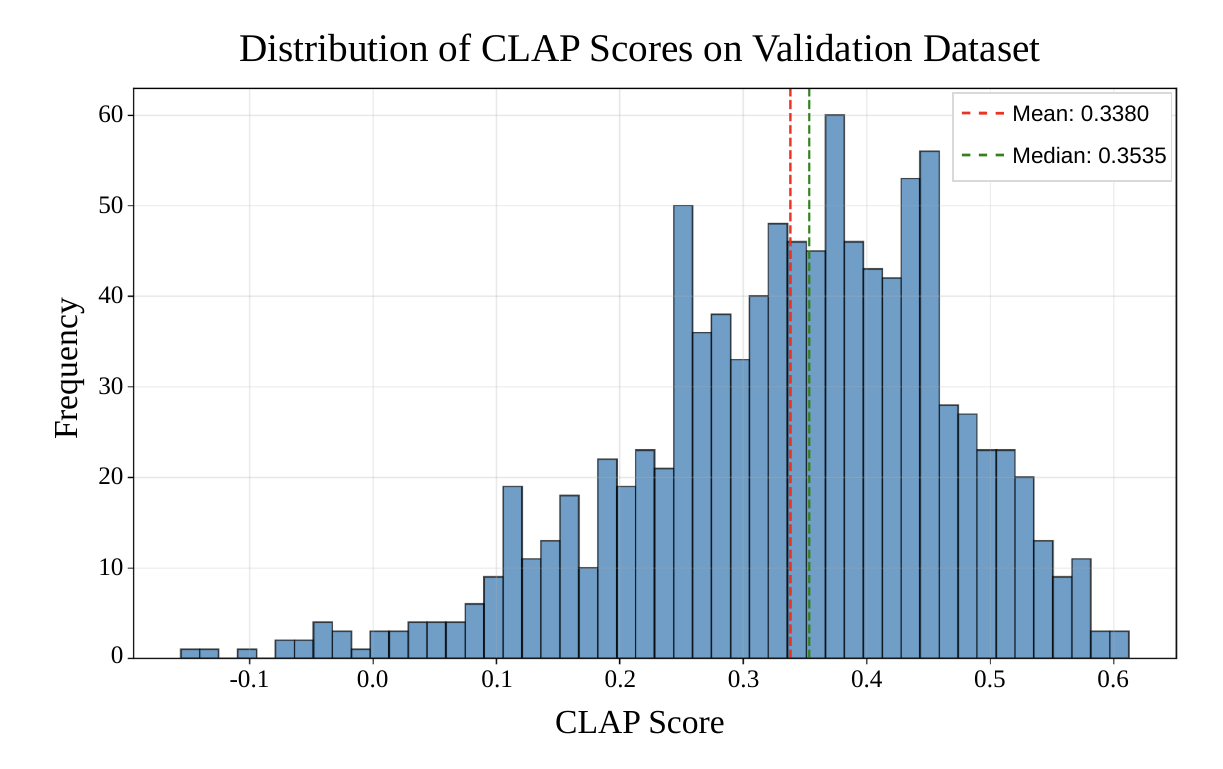}}
    \caption{CLAP score distribution of the validation dataset.}
    \label{fig:valClapDist}
\end{figure}

The first step in our score-aware training pipeline is to ensure that only sufficiently well-aligned audio-caption pairs enter the training pool. We begin by analyzing the CLAP score distribution of the validation dataset, computed on one randomly sampled 10-second segment from each of the 1,000 validation audio files. As shown in Fig.~\ref{fig:valClapDist}, the distribution has a mean of approximately 0.33, with a substantial proportion of segments exhibiting low CLAP scores. Moreover, since musical content can vary substantially within a single track (e.g., across verses, choruses, and instrumental sections), segments within the same audio file are likely to exhibit high variance in their CLAP scores, motivating a segment-level filtering strategy rather than file-level selection.

To ensure training data caption alignment quality, we apply a CLAP-guided segment selection pipeline. For each audio file, we randomly extract 15 candidate segments of 10 seconds each and compute the CLAP score for every segment. Segments are then partitioned into three quality tiers using two fixed thresholds: $\mathcal{S}_{\text{high}}$ (score $\geq 0.33$), $\mathcal{S}_{\text{medium}}$ ($0.20 \leq$ score $< 0.33$), and $\mathcal{S}_{\text{low}}$ (score $< 0.20$). Low-score segments are discarded entirely.

We aim to retain exactly 6 segments per file from the high and medium tiers, following a priority scheme:
\begin{enumerate}
    \item If the number of high-score segments $|\mathcal{S}_{\text{high}}| \geq 6$, we randomly sample 6 segments from $\mathcal{S}_{\text{high}}$.
    \item Otherwise, we take all $|\mathcal{S}_{\text{high}}|$ high-score segments and fill the remaining $6 - |\mathcal{S}_{\text{high}}|$ slots with the top medium-score segments.
\end{enumerate}

This strategy maximizes audio-caption alignment while maintaining sufficient training coverage per file.

\subsection{Pretraining and Caption-Aligned Fine-tuning}
\label{sec:data-preprocessing}

We adopt a two-stage training procedure designed to first build broad
musical knowledge from information-dense captions, and then specialize
the model to the concise prompt style encountered at inference time.

\paragraph{Stage 1: Pretraining on Information-Dense Captions.}
In the pretraining stage, we use both caption styles provided by the
ATTM challenge. \emph{Qwen-style} captions are generated directly by
Qwen2-Audio-7B-Instruct~\cite{chu2023qwenaudio}, producing holistic
descriptions of genre, instrumentation, and mood in a single pass.
\emph{MusicFlamingo-style} captions are generated by Music
Flamingo~\cite{ghosh2025musicflamingo} and subsequently refined by
Qwen3-4B-Instruct into concise, natural-sounding descriptions.
For each sample, one caption style is selected at random. Both styles are
information-dense and frequently include fine-grained musical attributes
such as tempo, key, time signature, and chord progressions, as illustrated
by the following example:
\begin{quote}
\textit{``Built on C$\sharp$ major in 4/4 time at 120 BPM, it follows a
loop-based structure with alternating chordal pads and melodic variations,
including diatonic and chromatic shifts, a brief minor passage, and a
resolved finale.''}
\end{quote}
This rich conditioning signal is well-suited for the early stages of
training: it provides a dense, structured supervision target that
encourages the model to learn fine-grained correspondences between
textual attributes and acoustic content.

\paragraph{Stage 2: Fine-tuning on Inference-Style Captions.}
While information-dense captions are valuable for pretraining, they
introduce a distributional mismatch with the prompts encountered at
inference time. Evaluation prompts tend to be concise and high-level,
focusing on genre, instrumentation, and mood---for example:
\begin{quote}
\textit{``An upbeat EDM track with pulsing synths and driving bass.''}
\end{quote}
A model trained exclusively on dense captions may underperform when
conditioned on these sparser prompts, simply because their distribution
was never seen during training. To bridge this gap, we introduce a
dedicated fine-tuning stage in which the model is adapted to the target
inference-time caption distribution.

We construct the fine-tuning corpus by rewriting each caption into the target style with a large language model using the following prompt%
\footnote{The \texttt{\{examples\}} placeholder is filled at runtime with
a small set of few-shot demonstrations sampled from the target evaluation
prompt distribution, anchoring the rewriter's output style to the desired
inference-time format.}:

\begin{tcolorbox}[
    colback=gray!8,
    colframe=gray!40,
    title=Caption Rewriting Prompt,
    fonttitle=\small\bfseries,
    fontupper=\small\ttfamily,
]
You are a music caption editor. Rewrite captions to include only:\\
1. Genre / style\\
2. Instrumentation (specific instruments)\\
3. Mood, theme, or atmosphere\\
\\
REMOVE: tempo/BPM, keys, time signatures, chord progressions, structural arcs, production/engineering descriptions.\\
\\
Write 1-2 natural flowing sentences. No bullet points. Vary the openings.\\
Match this style: \{examples\}\\
\\
Respond ONLY with a JSON array of rewritten captions in the same order. No preamble or markdown.
\end{tcolorbox}

The rewriting step strips low-level musical attributes (tempo, key,
chord progressions, structural arcs) while preserving the high-level
semantic content (genre, instrumentation, mood). Fine-tuning on these
rewritten captions teaches the model to ground generation in the same
sparse, high-level descriptors that users provide at inference, without
losing the fine-grained musical knowledge accumulated during pretraining.

\subsection{CLAP-Conditioned Noise Timestep Scheduling}
\label{sec:beta_schedule}

Rather than discarding segments whose alignment is imperfect but not negligible, our score-aware framework repurposes them by modulating their contribution to the training objective. Inspired by recent findings that low-quality data can still benefit generative training when used appropriately~\cite{li2025baddata}, we introduce a \textbf{CLAP-conditioned Beta noise timestep schedule} during flow matching. Instead of sampling the noise timestep $t \sim \mathcal{U}[0, 1]$ uniformly for all samples, we condition the timestep distribution on the CLAP score $S \in [0, 1]$ of each training segment using a Beta distribution:
\begin{equation}
    P(t \mid S) = \mathrm{Beta}\!\left(t;\, \alpha(S),\, \beta(S)\right),
\end{equation}
where $\beta(S) = 1$ and $\alpha(S)$ is a monotone function of $S$:
\begin{equation}
    \alpha(S) = 1 + \lambda\,(1 - S), \quad \lambda = 1.0.
\end{equation}
Here, $S$ is the CLAP score normalized to $[0, 1]$, and segments whose score exceeds the 75th percentile of the training distribution are treated as perfect quality, meaning their score is clipped to $S = 1$.

\begin{figure}[htbp]
    \centerline{\includegraphics[width=\linewidth]{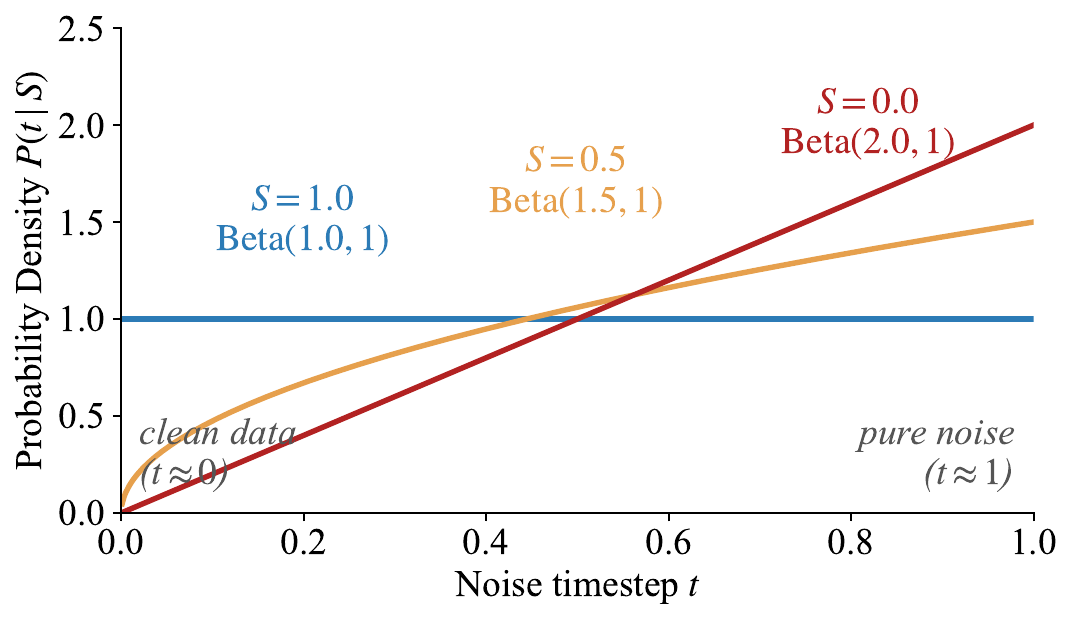}}
    \caption{Effect of the CLAP score $S$ on the timestep sampling distribution under the proposed Beta schedule ($\lambda=1.0$). High-score segments ($S = 1.0$) recover uniform sampling across all noise levels, while progressively lower-score segments concentrate their sampling mass near $t = 1$.}
    \label{fig:beta-distributions}
\end{figure}

The intuition relies on the observation that flow-matching models learn at different fidelities across the noise trajectory. For a high-score segment ($S \approx 1$), we obtain $\alpha \approx 1$, recovering the uniform distribution $\mathrm{Beta}(1, 1)$, which provides an equal training signal across all noise levels. Conversely, for a lower-score segment ($S \rightarrow 0$), we obtain $\alpha \approx 1 + \lambda$, which skews the Beta distribution toward $t \approx 1$ (the high-noise regime). At high $t$, predicting velocity only requires pointing in a roughly correct direction toward the clean data. By skewing the timestep sampling, lower-score segments contribute primarily to establishing this coarse semantic layout, without corrupting the fine-grained acoustic detail that is exclusively learned at low $t$. This targeted distribution of training signals is designed to act as an implicit regularizer, with the induced sampling behavior across score tiers summarized in Fig.~\ref{fig:beta-distributions}.

\subsection{REPA Alignment Loss}
\label{sec:repa}

\paragraph{Flow Matching Objective.}
Our backbone is trained with a conditional flow matching objective. Given a 
clean audio latent $\mathbf{x}_1$ and a noise sample $\mathbf{x}_0 \sim 
\mathcal{N}(\mathbf{0}, \mathbf{I})$, the noisy latent at timestep $t \in 
[0, 1]$ is constructed as $\mathbf{x}_t = (1 - t)\,\mathbf{x}_1 + 
t\,\mathbf{x}_0$, and the model is trained to predict the target velocity 
$\mathbf{v} = \mathbf{x}_0 - \mathbf{x}_1$ via mean squared error:
\begin{equation}
    \mathcal{L}_{\mathrm{FM}} = \left\| \hat{\mathbf{v}}_\theta(\mathbf{x}_t, t) - \mathbf{v} \right\|^2.
\end{equation}

\paragraph{Adding Representation Alignment.}
We augment this objective with a \textbf{Representation Alignment (REPA)}%
~\cite{yu2025repa} auxiliary loss, which encourages the model's hidden 
states (after the joint and fused transformer blocks) to align with 
structured embeddings from pretrained semantic encoders. We instantiate 
two branches: one targeting global audio-text semantics, the other 
fine-grained musical structure.

For the \textbf{CLAP branch}, hidden states are mean-pooled and projected 
by a trainable head $\phi_{\mathrm{CLAP}}$ into the CLAP embedding space, 
and the loss is the cosine distance to the CLAP embedding of the original 
audio:
\begin{equation}
    \mathcal{L}_{\mathrm{REPA\text{-}CLAP}} = 1 - \cos\!\left(\mathbf{z}_s,\; \mathbf{z}_{\mathrm{CLAP}}\right),
\end{equation}
where $\mathbf{z}_s = \phi_{\mathrm{CLAP}}\!\left(\tfrac{1}{T}\sum_{n=1}^{T} \mathbf{h}_n\right) \in \mathbb{R}^{d}$, $T$ denotes the audio sequence length (i.e., the number of frames rather than diffusion timesteps), $\mathbf{h}_n$ represents the hidden representation at the $n$-th frame, and $\mathbf{z}_{\mathrm{CLAP}} \in \mathbb{R}^{d}$ is the frozen CLAP embedding.

For the \textbf{MuQ branch}~\cite{zhu2025muq}, which captures music-specific 
structure such as timbre and instrumentation that CLAP's contrastive objective 
does not encode, we align at the \emph{sequence level}. Since both student 
hidden states and MuQ features operate at 25\,Hz, no resampling is needed: 
hidden states are projected frame-wise by a trainable head 
$\phi_{\mathrm{MuQ}}$, and the loss is the average cosine distance over 
all frames:
\begin{equation}
    \mathcal{L}_{\mathrm{REPA\text{-}MuQ}} = 1 - \frac{1}{T}\sum_{n=1}^{T} \cos\!\left(\mathbf{z}_{s,n},\; \mathbf{z}_{\mathrm{MuQ},n}\right),
\end{equation}
where $\mathbf{z}_{s,n} = \phi_{\mathrm{MuQ}}(\mathbf{h}_n) \in \mathbb{R}^{d'}$ and $\mathbf{z}_{\mathrm{MuQ},n} \in \mathbb{R}^{d'}$ is the corresponding frozen MuQ feature.

\paragraph{Timestep-Dependent Modulation.}
Both losses are modulated by $w(t) = (1-t)^{\alpha}$ with $\alpha = 2.0$, 
weighting alignment more heavily at low noise levels where the latent 
closely resembles the original audio. The full objective is:
\begin{equation}
\begin{split}
    \mathcal{L} = \mathcal{L}_{\mathrm{FM}} 
    &+ \lambda_{\mathrm{CLAP}} \cdot w_{\mathrm{CLAP}}(t) \cdot \mathcal{L}_{\mathrm{REPA\text{-}CLAP}}\\
    &+ \lambda_{\mathrm{MuQ}} \cdot w_{\mathrm{MuQ}}(t) \cdot \mathcal{L}_{\mathrm{REPA\text{-}MuQ}}.
\end{split}
\end{equation}

\section{Experiment}

\begin{figure*}[t]
    \centerline{\includegraphics[width=\textwidth]{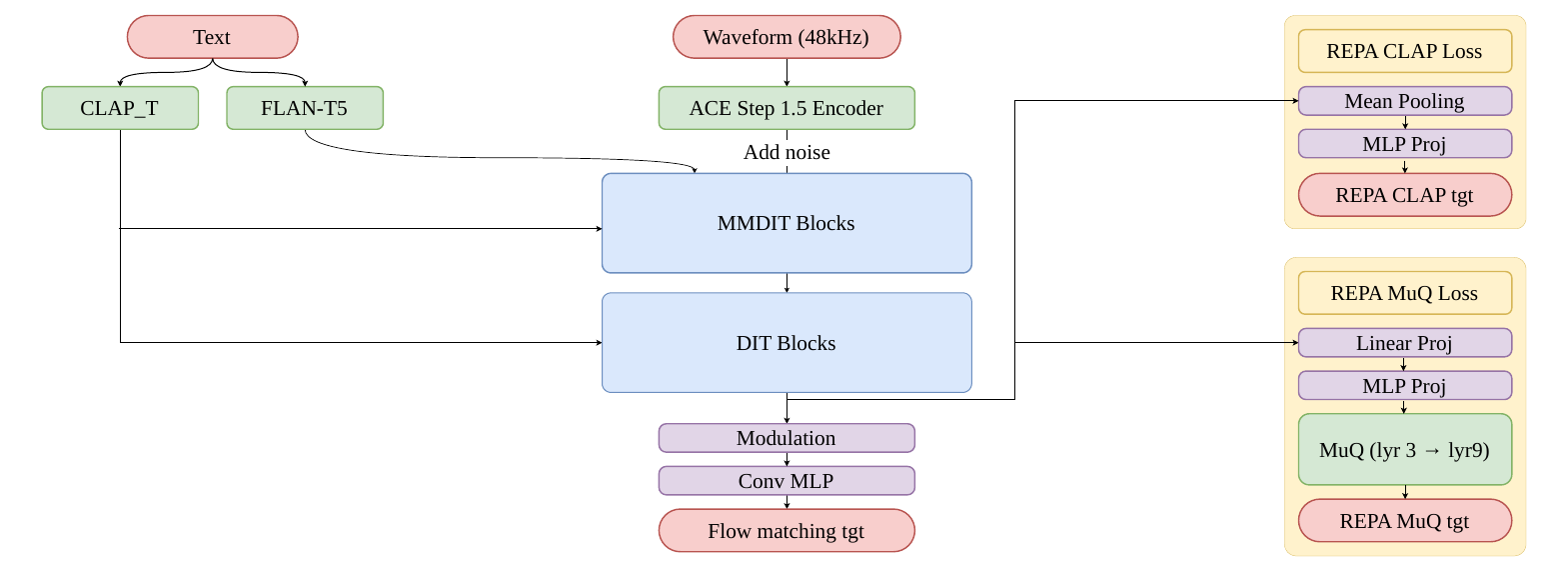}}
    \caption{Overview of our model architecture. The backbone is a FluxAudio Diffusion Transformer (DiT) conditioned on dual text representations: FLAN-T5 provides sequential token embeddings via cross-attention, while CLAP provides a global semantic embedding combined with the timestep embedding. Audio latents are encoded by the frozen ACEStep 1.5 codec. Hidden states extracted after the joint and fused transformer blocks are used for two auxiliary representation alignment branches: CLAP REPA (Setting 1 and 2) and MuQ REPA (Setting 2 only). At inference, the predicted flow velocity is decoded back to a waveform by the frozen ACEStep decoder.}
    \label{fig:architecture}
\end{figure*}

\subsection{Model Architecture}
We adopt \textbf{FluxAudio}~\cite{li2025meanaudio} as our flow matching backbone. FluxAudio is a FLUX-style~\cite{esser2024flux} Diffusion Transformer (DiT) trained with a conditional flow matching objective on audio latents. For the latent representation consumed by the flow matching process, we use the pretrained \textbf{ACEStep 1.5}~\cite{gong2026acestep15} audio codec as a frozen encoder, which encodes 48{,}000 Hz waveforms into continuous latent embeddings at 25 Hz. This provides compact, high-fidelity representations that preserve both acoustic detail and musical structure.

We condition the backbone on two complementary text representations: a \textbf{T5}~\cite{raffel2020t5} encoder provides fine-grained sequential token embeddings injected via cross-attention (sequence condition), while a \textbf{CLAP}~\cite{elizalde2023clap} encoder provides a global semantic embedding applied through adaptive layer normalization (global condition). For the REPA alignment branches, hidden states are extracted after the final DiT block: the CLAP branch mean-pools and projects these into the CLAP embedding space, while the MuQ branch projects and injects them into layers 3 through 9 of the frozen MuQ conformer. The overall architecture is illustrated in Fig.~\ref{fig:architecture}.

\subsection{Ablation Studies}

To isolate the contribution of individual design choices, we conduct controlled ablations on a reduced-scale setting: 2{,}000 training samples, 100 validation samples, trained for 20{,}000 iterations using the smaller \textbf{FluxAudio-S} backbone (hidden dim 448, depth 12, fused depth 8, 7 attention heads). The base configuration disables all optional components; each ablation activates exactly one component at a time. We report \textbf{CLAP score} (audio-text alignment, $\uparrow$) and \textbf{FAD} over CLAP embeddings ($\downarrow$) on the 100-sample validation set.

\begin{table*}[t]
\caption{Unified ablation results. The top row is the base configuration with all optional components disabled. Each subsequent group activates exactly one component while inheriting all other base settings (shown in gray). $\uparrow$ higher is better; $\downarrow$ lower is better.}
\label{tab:ablation-unified}
\begin{center}
\resizebox{\textwidth}{!}{%
\begin{tabular}{l l c c c c c}
\hline
\textbf{Ablation on} & \textbf{Configuration}
  & \textbf{CLAP REPA}
  & \textbf{MuQ REPA}
  & \textbf{Beta $\lambda$}
  & \textbf{CLAP} $\uparrow$
  & \textbf{FAD} $\downarrow$ \\
\hline
\textit{Base}
  & ---
  & \ding{55} & \ding{55} & 0
  & 0.2755 & 0.2856 \\
\hline
\multirow{2}{*}{CLAP REPA}
  & Normal ($\alpha{=}2.0$, $\lambda{=}0.2$)
  & \checkmark
  & \textcolor{gray}{\ding{55}} & \textcolor{gray}{0}
  & \textbf{0.2930} & 0.2767 \\
  & Aggressive ($\alpha{=}4.0$, $\lambda{=}0.4$)
  & \checkmark
  & \textcolor{gray}{\ding{55}} & \textcolor{gray}{0}
  & 0.2890 & \textbf{0.2620} \\
\hline
MuQ REPA
  & $\alpha=2.0, \lambda=0.1$
  & \textcolor{gray}{\ding{55}} & \checkmark
  & \textcolor{gray}{0}
  & 0.1921 & 0.5864 \\
\hline
\multirow{3}{*}{Beta Schedule}
  & $\lambda = 0.2$
  & \textcolor{gray}{\ding{55}} & \textcolor{gray}{\ding{55}}
  & 0.2
  & \textbf{0.2788} & 0.2941 \\
  & $\lambda = 1.0$
  & \textcolor{gray}{\ding{55}} & \textcolor{gray}{\ding{55}}
  & 1.0
  & 0.2746 & 0.2902 \\
  & $\lambda = 2.0$
  & \textcolor{gray}{\ding{55}} & \textcolor{gray}{\ding{55}}
  & 2.0
  & 0.2587 & 0.2995 \\
\hline
\end{tabular}}
\end{center}
\end{table*}

\paragraph{CLAP REPA}
Both REPA configurations improve over the base, with the normal setting yielding the best audio-text alignment ($+0.018$ CLAP score). The aggressive setting achieves the lowest FAD at a minor cost to CLAP score.

\begin{figure}[H]
    \centerline{\includegraphics[width=0.85\linewidth]{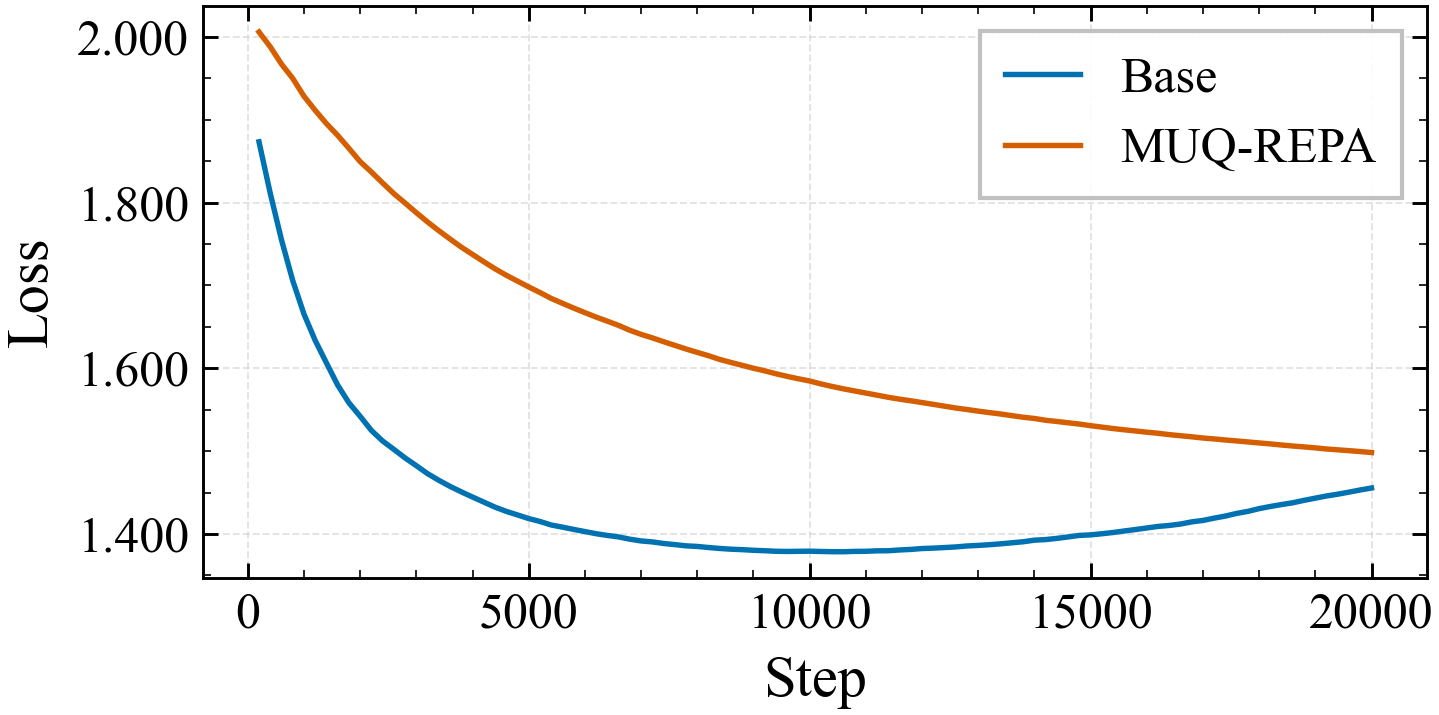}}
    \caption{Validation loss for base vs.\ MuQ REPA. The MuQ run starts from a higher initial loss (${\approx}2.0$ vs.\ ${\approx}1.88$) and converges substantially more slowly, confirming it has not converged within the 20{,}000-iteration budget.}
    \label{fig:muq-val-loss}
\end{figure}

\paragraph{MuQ REPA}
The MuQ alignment run shows substantial degradation in both metrics. We attribute this to two compounding issues: (1) as shown in Fig.~\ref{fig:muq-val-loss}, validation loss has not converged within 20{,}000 iterations, indicating that MuQ alignment requires a longer training horizon; and (2) CLAP score is an insufficient metric for the music-specific structure that MuQ encodes---timbre, instrumentation, and key are not reflected in text-audio contrastive alignment. 

\begin{figure}[htbp]
    \centering
    \includegraphics[width=0.85\linewidth]{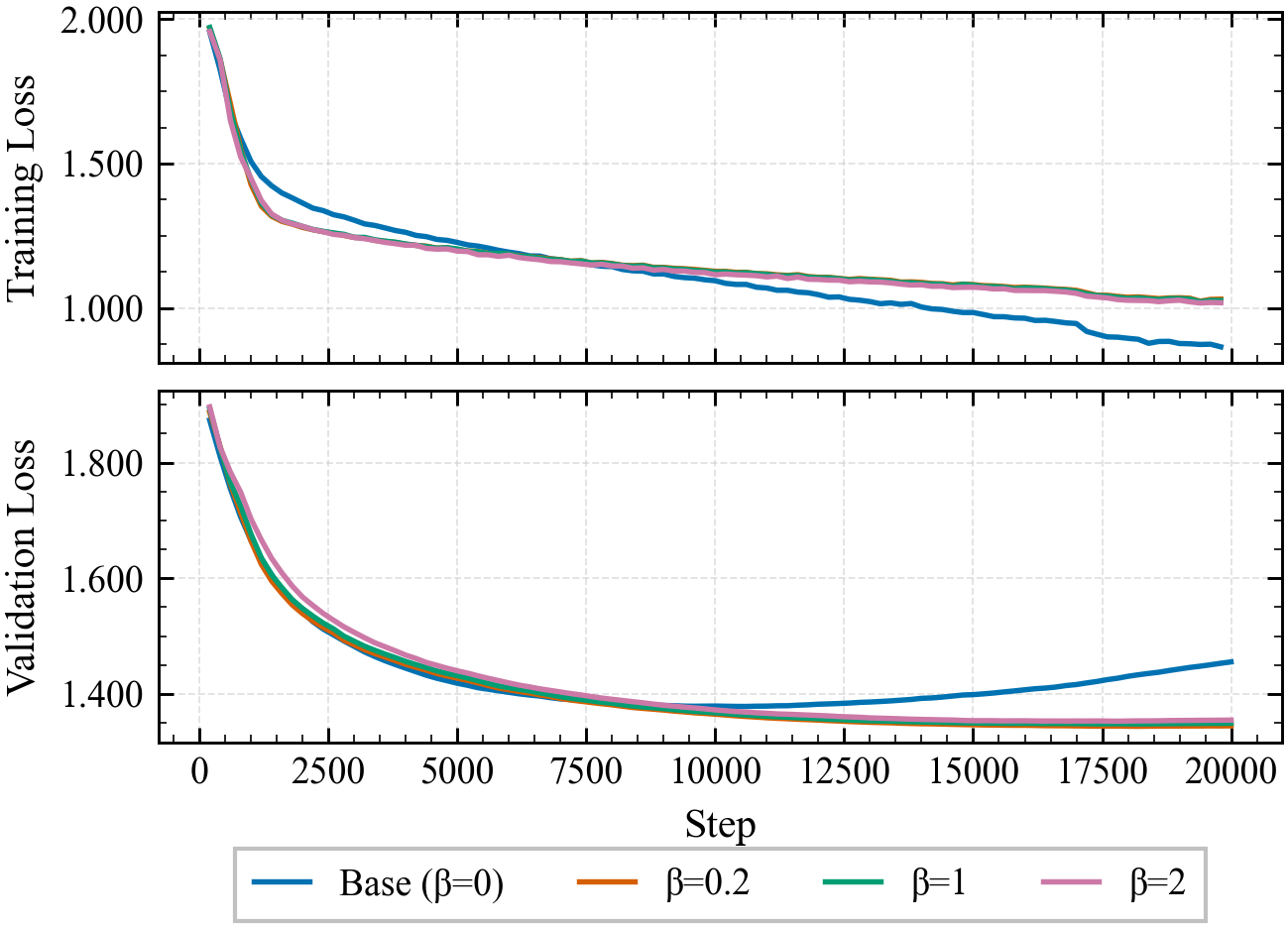}
    \caption{Training (top) and validation (bottom) loss for the Beta schedule
ablation. While the base model achieves the lowest training loss, it
overfits severely after step ${\approx}7{,}500$, with validation loss
rising sharply. All Beta variants cluster tightly in training and
generalize substantially better.}
    \label{fig:beta-loss}
\end{figure}

\paragraph{Beta Noise Schedule}
$\lambda = 0.2$ yields the best CLAP score ($+0.003$ over base), while $\lambda = 2.0$ degrades both metrics. The base achieves the best raw FAD, but this is misleading: as shown in Fig.~\ref{fig:beta-loss}, the base model overfits severely---validation loss reaches a minimum around step 7{,}500 before rising sharply to ${\approx}1.46$, while all Beta variants plateau at ${\approx}1.35$, a gap of 0.10 loss units.

Taken together, these ablations reveal a consistent theme: the primary challenge in this constrained setting is generalization rather than optimization. The base model overfits readily on the limited training set, as evidenced by the severe validation loss divergence in the Beta schedule ablation, demonstrating that the schedule acts as an effective implicit regularizer regardless of $\lambda$. CLAP REPA addresses this from a different angle, providing a semantically grounded auxiliary signal that stabilizes representation learning, yielding the most direct improvement in audio-text alignment ($+0.018$ CLAP score). The Beta noise schedule acts as a complementary implicit regularizer, reducing overfitting by approximately 0.10 validation loss units without sacrificing CLAP score. MuQ alignment, while theoretically well-motivated, requires a longer training horizon than our ablation budget permits and degrades both CLAP score and FAD within the 20{,}000-iteration budget.

\section{Submission Decision}

\subsection{Final Model Configuration}

Based on our ablation studies, we configure the final submitted model with approximately \textbf{450M} trainable parameters. Table~\ref{tab:hyperparams} summarizes the full architectural and training hyperparameters.

We retain \textbf{CLAP REPA} with the normal setting ($\alpha = 2.0$, $\lambda_{\mathrm{CLAP}} = 0.2$), as it consistently improves both CLAP score and FAD in controlled ablations. For the \textbf{Beta noise schedule}, we set $\lambda = 1.0$, which our ablations identify as stable and well-generalizing without over-aggressively suppressing training signal. We acknowledge that the ablations favor $\lambda = 0.2$; however, these results were finalized after the full training run had already been launched and could not be incorporated in time.

Given the uncertainty around MuQ alignment---which exhibited slow 
convergence within our ablation budget and could not be fairly assessed 
using CLAP-based metrics alone---we submit two variants: \textbf{Setting 1} 
(CLAP REPA + Beta) as our primary submission, and \textbf{Setting 2} 
(adds MuQ alignment) as an exploratory entry that did not fully converge 
before the deadline. Both settings undergo the caption rewrite fine-tuning 
stage described in Section~\ref{sec:data-preprocessing} (10{,}000 additional 
steps on simplified captions), which improves CLAP score from $0.304$ to 
$0.317$ ($+0.013$) on the final submission prompt set.

\begin{table}[htbp]
\caption{Architectural and training hyperparameters for the final submitted model.}
\label{tab:hyperparams}
\begin{center}
\begin{tabular}{lr}
\hline
\textbf{Parameter} & \textbf{Value} \\
\hline
\multicolumn{2}{l}{\textit{Architecture}} \\
Latent dimension              & 64 \\
Hidden dimension              & 896 \\
Transformer depth             & 12 \\
Fused depth                   & 10 \\
Number of attention heads     & 7 \\
Latent sequence length        & 250 \\
MLP ratio                     & 4.0 \\
Positional encoding           & RoPE \\
REPA projection dim           & 512 \\
MuQ projection dim            & 1{,}024 \\
Total trainable parameters    & ${\approx}450$M \\
\hline
\multicolumn{2}{l}{\textit{Training}} \\
Learning rate                 & $1 \times 10^{-4}$ \\
Weight decay                  & $1 \times 10^{-6}$ \\
Gradient clipping             & 1.0 \\
Linear warmup steps           & 1{,}000 \\
LR schedule                   & Step ($\gamma = 0.1$) \\
Mixed precision (AMP)         & Enabled \\
\hline
\multicolumn{2}{l}{\textit{CLAP REPA}} \\
$\lambda_{\mathrm{CLAP}}$               & 0.2 \\
$\alpha$ (timestep weight)              & 2.0 \\
\hline
\multicolumn{2}{l}{\textit{MuQ REPA}} \\
Enabled                                 & Setting 2 only \\
$\lambda_{\mathrm{MuQ}}$               & 0.1 \\
\hline
\multicolumn{2}{l}{\textit{Beta Noise Schedule}} \\
$\lambda$ (Beta skew)                   & 1.0 \\
$\beta(S)$                              & 1.0 \\
75th-percentile clip                    & $S \leftarrow 1.0$ \\
\hline
\multicolumn{2}{l}{\textit{Finetuning on Inference Style Captions}} \\
Steps                                   & 10{,}000 \\
Caption subset                          & 40\% of training data \\
\hline
\end{tabular}
\end{center}
\end{table}

\subsection{Objective Phase Results}
Our submission (Setting 1) achieves a CLAP score of $0.295$, FAD of $0.495$, and CCS of $0.804$ on the final test prompts, ranking \textbf{2nd across both tracks} in the objective evaluation phase and advancing as a finalist to the subsequent subjective listening test.

\subsection{Human Evaluation Phase Results}
Finalists advanced to a formal Mean Opinion Score (MOS) study conducted 
by expert listeners, evaluating \emph{Audio Quality}, \emph{Musicality}, 
and \emph{Prompt Adherence}. Our system (e08) achieved 
MOS$_{\mathrm{all}} = 3.119$ and MOS$_{\mathrm{expert}} = 3.044$, placing 
\textbf{3rd in the Efficiency Track}.

\section{Conclusion}

We presented a score-aware training framework for text-to-music generation 
under the constraints of the ICME 2026 ATTM Challenge. By treating caption 
alignment score as a first-class signal informing filtering, noise 
scheduling, caption preparation, and representation learning, we showed 
that careful handling of training dynamics can advance TTM in academic 
settings without industrial-scale data or compute. Our final 450M-parameter 
system ranked 2nd across both tracks in the objective phase and 3rd in the 
Efficiency Track in the expert MOS evaluation. Ablations on a smaller-scale 
training subset indicated that CLAP REPA improved audio-text alignment 
($+0.018$ CLAP score), the Beta noise schedule acted as a strong implicit 
regularizer (${\approx}0.10$ validation loss reduction), and caption 
rewriting helped bridge the training-inference distribution gap; these 
observations informed our final design, but full-scale verification is 
left to future work, along with optimizing the MuQ alignment objective 
over longer training horizons and extending the score-aware framework 
to other training signals beyond CLAP scores.

\section*{Acknowledgments}
The work is supported by grant from the Ministry of Education (MOE) of Taiwan (for Taiwan Centers of Excellence).
\bibliographystyle{IEEEbib}
\bibliography{references}

\end{document}